\documentclass{article} %
\usepackage{colm2024_conference}
\usepackage{algorithm}
\usepackage{colortbl}
\usepackage{algpseudocode}
\usepackage{algorithmicx}
\usepackage{microtype}
\usepackage{hyperref}
\usepackage{url}
\usepackage{booktabs}
\usepackage{amssymb}
\usepackage{wrapfig}
\usepackage{graphicx}
\usepackage{multirow}
\usepackage{subcaption}
\definecolor{darkblue}{rgb}{0, 0, 0.5}
\hypersetup{colorlinks=true, citecolor=darkblue, linkcolor=darkblue, urlcolor=darkblue}

\usepackage{amsmath,amsfonts,bm}

\def\eqref#1{equation~\ref{#1}}

\def\1{\bm{1}}

\def\vtheta{{\bm{\theta}}}

\def\mA{{\bm{A}}}
\def\mB{{\bm{B}}}

\def\mW{{\bm{W}}}
\def\mX{{\bm{X}}}

\def\mPhi{{\bm{\Phi}}}

\DeclareMathAlphabet{\mathsfit}{\encodingdefault}{\sfdefault}{m}{sl}
\SetMathAlphabet{\mathsfit}{bold}{\encodingdefault}{\sfdefault}{bx}{n}

\usepackage{lipsum}

\newcommand\blfootnote[1]{%
  \begingroup
  \renewcommand\thefootnote{}\footnote{#1}%
  \addtocounter{footnote}{-1}%
  \endgroup
}

\title{Personalized Collaborative Fine-Tuning \\ for On-Device Large Language Models}

\author{Nicolas Wagner, Dongyang Fan, Martin Jaggi \\
EPFL, Switzerland \\
\texttt{firstname.lastname@epfl.ch}
}

\colmfinalcopy %
\begin{document}

\maketitle

\begin{abstract}
We explore on-device self-supervised collaborative fine-tuning of large language models with limited local data availability. Taking inspiration from the collaborative learning community, we introduce three distinct trust-weighted gradient aggregation schemes: weight similarity-based, prediction similarity-based and validation performance-based. To minimize communication overhead, we integrate Low-Rank Adaptation (LoRA) and only exchange LoRA weight updates. Our protocols, driven by prediction and performance metrics, surpass both FedAvg and local fine-tuning methods, which is particularly evident in realistic scenarios with more diverse local data distributions. The results underscore the effectiveness of our approach in addressing heterogeneity and scarcity within local datasets.
\end{abstract}

\section{Introduction}

Recently, there has been an unprecedented surge in the popularity of Large Language Models (LLMs), driven by their versatile capability to solve a wide range of tasks and serve as general-purpose models~\citep{touvron2023llama,openai2024gpt4,jiang2024mixtral}. Their downstream performances can be further enhanced by fine-tuning, which is typically conducted on fewer data and independently by different users, rather than in a centralized way. The scarcity of local data often renders local fine-tuning ineffective, necessitating collaboration. However, in many cases, end users usually have privacy concerns over their local data, such as patient records for hospitals~\citep{li-etal-2023-two} and typing history for mobile phone users~\citep{gboard}. Naturally, one might wonder \emph{how we could enable users to collaborate to still obtain better models in the presence of small and privacy-sensitive local data.}\blfootnote{Our codes are available at \url{https://github.com/epfml/personalized-collaborative-llms}}

The possibility of on-device fine-tuning of LLMs has been facilitated by parameter-efficient techniques such as prompt tuning~\citep{lester2021power}, adapters~\citep{houlsby2019parameter} and Low-Rank Adaptation (LoRA,~\cite{hu2022lora}). LoRA has gained widespread popularity for fine-tuning large language models since its introduction, which approximates weight updates through the multiplication of low-rank matrices. LoRA modules align with our collaboration initiative, as they allow for a significant reduction in communication overhead. A naive approach to leveraging local data from other users would be to average LoRA weights, which is the popular FedAvg algorithm proposed by \citet{mcmahan2017communication}. However, the resulting one global model solution might not fit all users' data distributions, especially when there is more heterogeneity present. As a result, we aim to develop personalized collaboration strategies for each user, which should surpass the performance of local fine-tuning and naive averaging approaches.

In pursuit of this objective, we introduce several collaborator selection protocols tailored for optimal LoRA weight averaging across diverse user profiles. While similar concepts have been extensively investigated within collaborative learning communities~\citep{zhang2021personalized, sui2022friends, fan2024collaborative}, these approaches are only designed for supervised classification tasks and thus are not directly applicable to Large Language Models (LLMs). Nevertheless, we investigate how we can adapt and apply analogous principles to \emph{determine optimal aggregation weights for LoRA matrices on a per-user basis}. 

Following some computational social science literatures~\citep{urena2019review, zhang2022consensus}, we term the aggregation weight matrix ``trust'', as if we view clients as members of a social group, their exchanged information can thus be regarded as opinions. Similarity measures can be useful for assessing trust between users, as users are more likely to consider and trust others with similar opinions. Such similarity measures are core to our proposed protocols, which we will discuss in Section~\ref{sec: protocols}.

Our contributions can be summarized in the following aspects:
\begin{itemize}
    \item For the first time in the language domain, we explore realistic data heterogeneity among users, which involves diverse topic distributions and varying language usage. This introduces a new challenge as well as use case for language modeling. 
    \item We demonstrate that collaboration can be leveraged to improve personalization performance in language modeling and present 3 unique collaboration protocols for identifying optimal collaborators.
    \item Our approach is well-suited for on-device fine-tuning, effectively mitigating challenges related to data scarcity and resource limitations.
\end{itemize}

\section{Related Works}

We focus on a \emph{peer-to-peer decentralized learning} setting, where the existence of a central server is not assumed. Instead, the end users conduct peer-to-peer communication using decentralized schemes such as gossip averaging to aggregate local information across agents. To enable a personalized weighted aggregation, several recent works have proposed data-dependent communication protocols based on task similarities and node qualities. Notably, \cite{zhang2021personalized} derives a first-order approximation for optimal aggregation weights $w_{ij}^\star$, which happens to be proportional to how well client $j$'s model generalizes on client $i$'s data. \citet{li-l2c2022} directly optimizes the mixing weights by minimizing the local validation loss per node. \citet{sui2022friends} uses the E-step of the EM algorithm to gauge the significance of other agents for a specific agent $i$, achieved through assessing the accuracy of these agents' models on the local data of agent $i$. \citet{fan2024collaborative} adopt a different approach by comparing prediction similarity on a publicly available dataset, which largely reduces communication overhead and allows for model heterogeneity. These approaches have been showing promising performances in supervised deep learning experiments. In the language domain, the data heterogeneity across users is not well studied, and the effectiveness of the collaborator selection methods has not been explored for LLMs, where self-supervised training is performed.

Recently, groups of researchers have been investigating the intersection of Federated Learning and Large Language Model training. Due to the substantial number of model parameters, these efforts primarily focus on the post pre-training stages, often incorporating parameter-efficient techniques. We refer to the work of \citet{fan2023fatellm} for a detailed framework. Among the endeavors,
\cite{zhang2023towards} employ FedAvg for collaborative instruction tuning, where they demonstrate superior performance in task generalization compared to local instruction tuning. \citet{che-etal-2023-federated} design an adaptive optimization
method for collaborative prompt tuning, where a scoring method is applied to measure the importance of each layer and only prompt parameters of more important layers are exchanged during the tuning process, ensuring communication efficiency. \citet{cho2024heterogeneous} tackles the challenge brought up by different LoRA ranks on heterogeneous devices. By performing rank self-pruning locally and sparsity-weighted aggregation at the server, improved convergence speed and final performance can be achieved. In contrast to these methods, we aim at providing personalized models to each user, instead of a global solution. 

\section{Methods}

To start with, we first introduce how LoRA works. For a pre-trained model parameterized with $\mPhi \in \mathbb{R}^{m\times n}$, the model updates during fine-tuning stage can be approximated by multiplication of two low-rank matrices $\mA \in \mathbb{R}^{m\times r}$, $\mB \in \mathbb{R}^{r\times n}$
\begin{equation} 
    \Delta \mPhi \approx \mA \mB, \qquad \text{where}\qquad r \ll \min(m,n)
\end{equation}

We work in a setting where all users have the same LoRA rank $r$, and aim to arrive at their personalized $\Delta \mPhi_i$. During the fine-tuning phase, only LoRA parameters will be updated, and the original weights of the pre-trained model will remain frozen. Therefore, it suffices to communicate LoRA weights, instead of the big chunk of whole parameters.

\subsection{Our Protocol}
\label{sec: protocols}
\begin{figure}
\vspace{-2em}
    \centering
    \includegraphics[width=.7\textwidth]{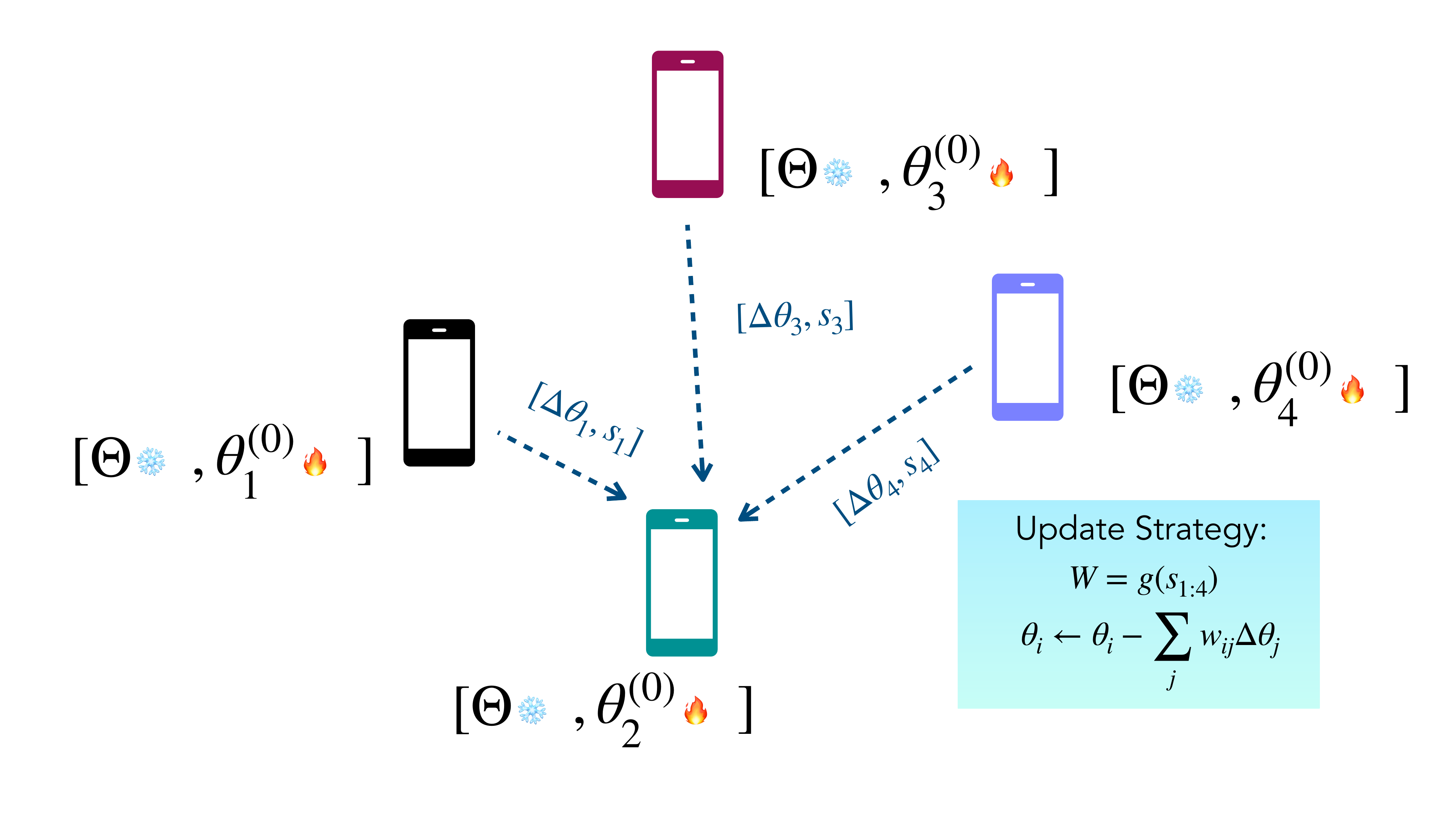}
    \caption{Diagram of our protocol. $\vtheta_i$ and $\Delta \vtheta_i$ represent LoRA weights and LoRA weight updates respectively. $s_i$ denotes messages to send beside $\Delta \vtheta_i$, which represents either $\vtheta_i$ or $f_{\vtheta_i}(\mX_S)$ depending on the protocol (see Table~\ref{tab: communication_vs_computation}). $g(\cdot)$ is our proposed trust calculation approach as detailed in Section~\ref{trust-calculation}.}
    \label{fig:diagram}
\end{figure}

We stick to the standard approach of sharing and aggregating model updates. Each user, through their unique data distribution, contributes to the collective learning process by sharing local model updates. Central to our approach is the so-called \emph{trust matrix}, which we define as the gossip aggregation matrix to guide each user's gradient aggregation. The calculation of trust matrix $\mW$ will be detailed in Section~\ref{trust-calculation}. A simplified diagram of our protocol is shown in Figure~\ref{fig:diagram}.

Our proposed protocol is described in Algorithm \ref{alg:gradient}. To summarize, the protocol is constituted with three phases:  1) each user $i\in [N]$ computes their LoRA weight update $\Delta \vtheta_i$ with respect to their local data (and if required, predictions $f_{\vtheta_i}(\mX_S)$ with respect to the shared dataset $\mX_S$); 2) each user communicates $[\vtheta_i, \Delta \vtheta_i]$ or $[f_{\vtheta_i}(\mX_S), \Delta \vtheta_i]$ to all other users, depending on the trust calculation strategy; 3) each user $i \in [N]$ calculates their unique trust weights $\mW_{i\cdot}$ locally in parallel and updates their LoRA weights using trust-weighted received gradients. It is noteworthy that different means of trust weight calculation require different information to be shared across the users, thus inducing distinct computation complexities and communication overheads. We will discuss those aspects in Section~\ref{remarks}.

\begin{algorithm}[t!]
\caption{Our proposed protocol}\label{alg:gradient}
\begin{algorithmic}
\Require Number of communication rounds T, pre-trained model $\bm{\Theta}$, initialized LoRA weights $\vtheta_i^{(0)}$, local dataset $\mathbf{X}_i$, shared dataset $\mathbf{X}_S$ (required for strategy~\ref{strategy3}).\\

\For{$t = 1, ...,T$}
\State 1. \textbf{in parallel} for each user $i \in [N]$ \textbf{do}
\State \indent Compute update $\Delta \vtheta_i^{(t - 1)}$ on $\vtheta_i^{(t - 1)}$ with respect to $\mathbf{X}_i$
\State 2. Each user $i \in [N]$ broadcast their $\Delta \vtheta_i^{(t - 1)}$ to all others agents
\State 3. \textbf{in parallel} for each user $i \in [N]$ \textbf{do}
\State \indent  Calculate the trust matrix $\mathbf{W}$, based on a defined strategy (\ref{strategy1}, \ref{strategy2} or \ref{strategy3})
\State \indent  Compute trust-weighed personalized weight update: $$\vartheta_i^{(t)} = \sum_{j \in [N]} w_{ij} \Delta \vtheta_{j}^{(t - 1)}$$
\State \indent  Update LoRA parameters, with learning rate $\eta$ $$\vtheta_i^{(t)} = \vtheta_i^{(t - 1)} - \eta \vartheta_i^{(t)}$$
\EndFor
\end{algorithmic}
\end{algorithm}

\subsection{Trust Calculation}
\label{trust-calculation}
In this section, we explore different ways of building trust -- gossip aggregation graph across users. Essentially, the weight of edge $(i,j)$ should denote to what extent can user $j$'s gradients help to facilitate user $i$'s learning progress.

\subsubsection{Weights similarity based}
\label{strategy1}
We employ pairwise LoRA weight similarity to compute trust scores among users. The underlying idea is that users whose model weights exhibit closer alignment likely possess data distributions that are more similar, thereby rendering their contributions more pertinent and advantageous to one another. This methodology shares a resemblance with \cite{li-l2c2022}, where the mixing weight is derived from the inner product between representations of different users' models via an encoder model. However, our approach involves directly measuring the similarity between model updates, without the need for an encoding module.

It appears that in addition to updating LoRA weights, they also need to be shared to compute the trust matrix. However, this additional communication can be efficiently circumvented, as the updates to LoRA weights are already exchanged, and the weights themselves can be readily derived by aggregating these updates in each iteration. The trust matrix $\mathbf{W}$ is computed as follows:
\begin{equation}
\begin{aligned}
     \tilde{w}_{ij}^t = \frac{\langle \vtheta_i^{(t - 1)}, \vtheta_j^{(t - 1)} \rangle}{\|\vtheta_i^{(t - 1)}\|\|\vtheta_j^{(t - 1)}\|}, \qquad
     \mW^t = \text{SoftMax} (\widetilde{\mW}^t, \text{dim}=1)
\end{aligned}
\end{equation}

\subsubsection{Validation performance based}
\label{strategy2}
Drawing inspiration from \citet{zhang2021personalized}, we evaluate the pairwise trust $w_{ij}$ based on how well user $j$'s model performs on the validation sets of user $i$. This method requires extra validation sets within each user, on which the performance assessment directly reflects the relevance and potential benefit of one user's model to another. The aim is to guide the aggregation process by favoring models that demonstrate compatible performance levels, thereby enhancing the overall effectiveness of collaborative learning. 
The calculation is denoted as:

\begin{equation}
    \begin{aligned}
        \tilde{w}_{ij}^t = \mathcal{L}(f_{\vtheta_j^{t-1}}(\mX_i^{\text{val}}), \mX_i^{\text{val}}), \qquad \mW^t = \text{SoftMax} (-\widetilde{\mW}^t, \text{dim}=1)
    \end{aligned}
\end{equation}

where $f_{\vtheta_j^{t-1}}(\mX_i^{\text{val}})$ denotes a forward pass of user $j$'s model on user $i$'s local validation set, and $\mathcal{L}$ is the cross entropy loss for next token prediction.

\subsubsection{Prediction similarity based}
\label{strategy3}
In the prediction similarity-based approach, following the trust weighing strategy from \citet{fan2024collaborative}, we derive aggregation weights by comparing the model's predictions on a shared dataset, in this case, logits across the vocabulary. A public dataset $\mX_S$ will be added to the input of Alg.~\ref{alg:gradient}, and additionally communicating those predictions between users is necessary, increasing communication cost. 
This method assesses how closely the predictions of different models align, prioritizing contributions from users whose models exhibit similar predictive behaviors. The trust is calculated based on $l_1$ distance between logits:
\begin{equation}
    \begin{aligned}
        \tilde{w}_{ij}^t = |f_{\vtheta_i^{t-1}}(\mX_S) - f_{\vtheta_j^{t-1}}(\mX_S)|, \qquad
        \mW^t = \text{SoftMax} (-\widetilde{\mW}^t, \text{dim}=1)
    \end{aligned}
\end{equation}

\subsection{Remarks}
\label{remarks}
Table~\ref{tab: communication_vs_computation} provides an overview of the communication and computation costs associated with the three proposed strategies.
\paragraph{On communication costs} In terms of communication, both strategies 1 and 2 need to share LoRA weights $(\vtheta_i)$ on top of LoRA weight updates $(\Delta \vtheta_i)$.  Strategy 3 requires sharing logits ($f_{\vtheta_i}(\mX_S)$) across the public dataset, rather than LoRA weights. Although the logits encompass the entire vocabulary, it is possible to heavily reduce its memory cost by artificially transforming it into a sparse matrix. The computation of trust through prediction similarity relies on the $l_1$ distance, and most entries of logits tend to be very close to zero, making it feasible to only keep the top-k logits values. This approach greatly cuts down on communication costs when sharing logits while still maintaining desired performance levels. For a detailed comparison of the measured communication costs, please refer to Appendix~\ref{tab: communication costs in mb}. Nonetheless, as the communication of $\Delta \vtheta_i$s remains unavoidable and constitutes the primary communication cost, there is not much variation in communication costs across all three strategies.

\paragraph{On extra computation costs} 
Regarding additional computation costs, Strategy 1 involves only calculating cosine similarities between LoRA weights in addition to existing processes, the cost of which is in the order of the number of LoRA parameters. However, Strategies 2 and 3 necessitate multiple forward passes to generate predictions either on local validation sets or the publicly shared dataset. A forward pass is significantly more resource-intensive compared to cosine similarity calculations due to the model's size. In Strategy 2, evaluating all models' performance on each local validation set requires 
$N^2$ model inferences. In Strategy 3, each user needs extra forward passes to obtain predictions on $\mX_S$, resulting in $N$ times the model inference. Since $\mX_S$ and $\mX_i^\text{val}$ are similar in size, the number of forward passes per inference doesn't vary much. Thus, strategy 2 requires roughly $N$ times more extra computation cost than strategy 3.

\begin{table}[h!]
\begin{center}
\begin{tabular}{p{3cm} | p{4cm} | p{5cm}}
\toprule
\bf Strategy  & \bf Communicated Elements & \bf Extra Computation Costs\\
\midrule
1: Weights         & $\{\vtheta_i, \Delta \vtheta_i\}_{i=1}^N$  & $\mathcal{O}(Ldk)$ \hfill $\star$\\
2: Validation             &$\{\vtheta_i, \Delta \vtheta_i\}_{i=1}^N$ & $\mathcal{O}(Ln^2d + Ld^2n)\cdot N^2 $ \hfill $\star \star \star$\\
3: Predictions             &$\{f_{\vtheta_i}(\mX_S), \Delta \vtheta_i\}_{i=1}^N$ & $\mathcal{O}(Ln^2d + Ld^2n)\cdot N $ \hfill $\star \star$\\
\bottomrule
\end{tabular}
\end{center}
\caption{A comparison of communication and computation complexity across the strategies. $L$ denotes the number of layers, $d$ denotes the embedding size, $n$ denotes context length and $k$ denotes LoRA rank. $N$ denotes the number of users in the system. It is clear strategy 2 adds the most extra computational cost, followed by strategy 3. Strategy 1 requires the least amount of computation on top.}
\label{tab: communication_vs_computation}
\end{table}

\section{Experiments}
\subsection{Setup}

\subsubsection{Experimental details}
Our experimental setup was structured as follows: Each user was configured with a uniform model architecture, to ensure model updates can be shared and aggregated. We chose relatively small LLMs constrained by the limited computing resources in academia. For each user, we equip them with a GPT2\footnote{https://github.com/karpathy/nanoGPT} base model, with 124 million parameters in total. It has 12 layers, 12 attention heads, embedding size 768 and vocabulary size 50257.

\begin{wrapfigure}{tr}{0.45\textwidth}
  \begin{center}
\includegraphics[width=0.45\textwidth]{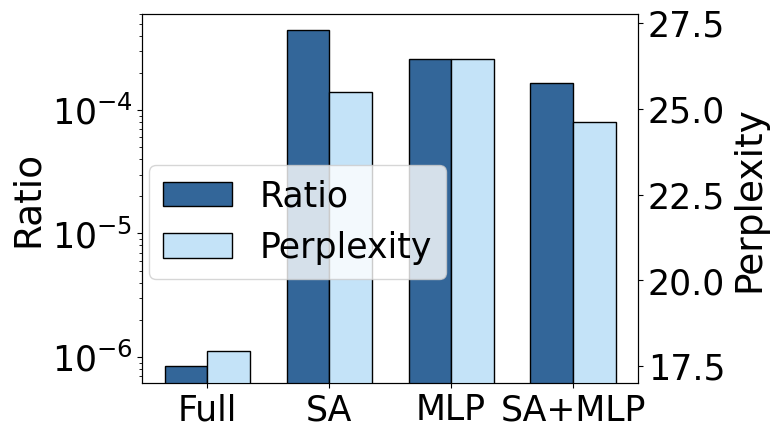}
  \end{center}
  \caption{Ablation study of whether to add LoRA modules compared to full fine-tuning. Left bars correspond to \emph{Ratio}, which denotes the decrease in test perplexity compared to the pre-trained model per trainable parameter (higher is better); right bars correspond to \emph{perplexity}, which denotes test performance after fine-tuning (lower is better).}
  \label{fig: barplot-lora}
  \vspace{-1em}
\end{wrapfigure}

LoRA modules were applied on top of both Causal Self-Attention (SA) and MLP layers, constituting $0.47\%$ of full parameters. The reason why we chose the two blocks is a trade-off between the model performance and communication overhead, as shown in Figure~\ref{fig: barplot-lora}. Although incorporating LoRA modules onto Self Attention layers results in the most significant reduction in test perplexity per trainable parameter, integrating LoRA modules into MLP layers can contribute to further decreasing test perplexity while maintaining a reasonable number of trainable parameters.

The experiments were conducted, for each client, with a learning rate of 0.002, a batch size of 50 with 4 accumulation steps, a context length of 512, and a total of 500 iterations. Every user performs local training for 100 iterations before communication, so that the local models can start to diverge due to the variation of local data distributions. Afterward, communication and aggregation are done every 25 iterations. As we target a limited data regime, the number of iterations ensures that over 50 epochs of training are conducted in most datasets. Throughout all experiments, we use the same LoRA configurations: a rank of 4, an alpha value of $32.0$, and a dropout rate of 0.1. The selection of hyperparameters is determined through a sweep of various values while considering the limitations posed by our computing resources, as presented in Appendix~\ref{append :hyperparameters}.

\subsubsection{Datasets}
In vision tasks, realistic data heterogeneity arises from diverse domain distributions across clients and variations in image quality resulting from the use of different devices for data acquisition~\citep{ogier2022flamby}. However, within the domain of written language, realistic data heterogeneity frequently arises from variations between sources. Owners of such sources can differ in writing styles, vocabulary choices, grammar structures, and topic distribution. Our experiments are designed to examine diverse topic distributions and language usages exhibited by different users.

We focus on on-device fine-tuning scenarios, where the amount of available local training data may be limited. The diverse local data distribution originates from user-specific characteristics. For instance, users often possess unique topic preferences concerning news reading, and mobile phone users often type in different languages, prompting smart keyboards to enhance next-word prediction across different language blends. Following this, we investigate two levels of data heterogeneity: 1) \emph{Low heterogeneity}, where each user is assigned two data categories from the entire set, and a $(3/4,1/4)$ mixture of these categories is allocated to each user. 2) \emph{High heterogeneity}, where each user is exclusively assigned to one data category.

The descriptions of the three dataset partitionings utilized in our experiments are as follows: 
\begin{enumerate}
    \item AG News: News articles in four categories: ``World'', ``Sports'', ``Business'' and ``Sci/Tech''.~\citep{zhang2015character}
    \item Multilingual Wikipedia: Wikipedia texts in three languages (categories): French, Italian, and German.~\citep{wikidump}
    \item Codes-Wikipedia (Eng): The first category is Java code from GitHub~\citep{github-ds} and the second category is English Wikipedia text~\citep{wikidump}.
\end{enumerate}

The corresponding distributed datasets are presented in Table~\ref{table:dataset_config}. Note that AG News and Codes-Wikipedia simulate different topic distributions across users, while Multilingual Wikipedia simulates different language usage across users. It is worth noting that even though the topics/languages are different, they can nevertheless share the same tokens in their vocabularies. The portion of overlapped tokens is indeed not small, as measured by the Jaccard Index in Table~\ref{table:jaccard}. Jaccard Index for multi-sets, is calculated as $J(\mathcal{D}_i, \mathcal{D}_j) = \frac{\lvert\mathcal{D}_i \cap \mathcal{D}_j\rvert}{\lvert\mathcal{D}_i\rvert + \lvert\mathcal{D}_j\rvert} \in [0,0.5]$. A high Jaccard Index indicates that clients' data share a substantial number of the same tokens. Although this measure doesn't account for the context information, it provides insight into the likelihood of finding collaborators. In all our experiments, the shared public dataset $\mX_S$ is sampled equally from each local distribution to ensure that the logit predictions are meaningful. In practice, such $\mX_S$ can be chosen as any publicly available dataset from the same domain, or private data with data sanitization procedures enforced on top to secure user privacy.

\begin{table}
\begin{center}

\begin{tabular}{c| c c c c} 
 \toprule
 \bf Datasets & \bf \# users & \bf \# Categories & \bf \# Training tokens & \bf \# Test tokens \\ [0.2ex]
  \midrule
 AG News & 4/8 & 4 & $\sim1'500'000$/$\sim750'000$ & $\sim75'000$/$\sim50'000$ \\[0.2ex]
 Multi-Wiki & 9 & 3 & 840'000 & 160'000 \\[0.2ex]
 Codes-Wiki & 8 & 2 & 840'000 & 160'000 \\[0.2ex]
 \bottomrule
\end{tabular}
\end{center}
\vskip -0.1in
\caption{Dataset configurations used in the experiments. Reference data $\mX_S$ is slightly larger in size than test tokens and $\sim100'000$ tokens are sampled for logits computation. There is insufficient data from AG News to establish a meaningful public reference dataset, so we resort to ChatGPT to generate a synthetic one.}
\label{table:dataset_config}
\end{table}

\begin{table}
\vskip 0.15in
\begin{center}
\begin{tabular}{ c c | c c } 
 \toprule
 \bf Datasets & \bf Heterogeneity & \bf with the most task overlap & \bf with the least task overlap \\ [0.2ex]
 \midrule
 \multirow{2}{*}{AG News} & Low & 0.39-0.41 &  0.31-0.32\\
 & High & 0.42-0.44 & 0.28-0.33 \\ \midrule
 \multirow{2}{*}{Multi-Wiki} & Low & 0.45-0.46 & 0.19-0.24\\
 & High & 0.45-0.47 & 0.19-0.23  \\ \midrule
 \multirow{2}{*}{Codes-Wiki} & Low & 0.40-0.41 & 0.39-0.42\\ 
 & High & 0.41-0.42 & 0.12-0.13  \\ 
 \bottomrule
\end{tabular}
\end{center}
\vskip -0.1in
\caption{Jaccard Index values for various datasets, indicating the range of token overlap among clients in different configurations.}
\label{table:jaccard}
\vspace{-1em}
\end{table}

\subsubsection{Baselines}
Our chosen baseline methods are: 1) Local Fine-Tuning, where there is no communication between users, 2) FedAvg~\citep{mcmahan2017communication}, where model updates are aggregated to obtain a global model for each user, and 3) FedAvg + Local Fine-Tuning~\citep{jiang2023improvingfederatedlearningpersonalization}, where we spare 10\% of training data for local fine-tuning after FedAvg training. We offer an extended table where we vary the amount of training data used for fine-tuning for completeness in Table~\ref{app:table:res_summary} in Appendix. We further incorporate a strong baseline -- \emph{oracle}.  The determination of \emph{oracle} collaboration weights relies on the similarity of underlying data source distributions, which is typically \emph{unknown} in practice. We give an example of how the \emph{oracle} weights are determined: if user 1 has 1/4 German texts and 3/4 French texts, and user 2 has 1/4 German texts and 3/4 Italian texts, the oracle weights would be determined by the dot product of $[1/4,3/4,0]$ and $[1/4,0,3/4]$. After obtaining the pairwise dot product matrix, row normalization is performed.

\subsection{Results}

Our main findings are summarized in Table~\ref{table:res_summary}. Compared to baseline approaches—local training, FedAvg, and FedAvg + FT—our test performance and prediction-based aggregation methods consistently outperform them across all datasets and scenarios. We emphasize that in language modeling, achieving a balance between personalization and collaboration is not straightforward. Despite significant differences between categories, there are notable similarities in sentence structure and word choices, alongside domain-specific vocabulary. The collaboration-then-personalization approach (FedAvg + Local Fine-Tuning) fails to achieve this balance. However, our methods successfully accomplish it.

Among our protocols, the predictions-based method exhibits the most superior test performance. Notably, the weights similarity-based aggregation method yields performance akin to FedAvg, where the trust weight is uniformly set to $1/N$ across all clients. Our methods can sometimes outperform \emph{oracle} aggregation, suggesting that a \emph{dynamic} collaborator selection protocol might be favored in different fine-tuning stages.

Ablation studies on the effect of LoRA ranks and local training set size can be found in Table~\ref{app: tab-lora-rank} and Table~\ref{app: tab-local-dataset-size} in Appendix~\ref{append: abaltions}. Our prediction-based strategy consistently delivers superior performance.

\begin{figure}[t!]
\begin{minipage}{0.55\textwidth}
    \centering
    \begin{tabular}{l|c c | c c}
\toprule
\multirow{2}{*}{Method} & \multicolumn{2}{c}{\textbf{Multi-Wiki}} & \multicolumn{2}{c}{\textbf{Codes-Wiki}} \\
 &	 Low	& High& 	 Low & High \\
\midrule
 Local	& 1 & 	1	& 1	 & 1\\
FedAvg& 	1.6	& 4.2	& 2.0& 	1.7 \\
Strategy 1	& 5.9	& 12.0	& 2.0	& 2.1 \\
Strategy 2	& 1.1	& 2.6	& 1.5& 	NA \\
Strategy 3	& 2.9	& NA	& 1.2	& 3.8 \\
\bottomrule
    \end{tabular}
    \captionof{table}{ Training time ($X$ times the needed training iterations as in the fully connected case) required for a ring topology to achieve the same perplexity level as a fully connected topology. NA indicates we did not reach the same perplexity after ten times the training iterations.}
    \label{tab:ring-topology}
    \end{minipage}
    \hfill
    \begin{minipage}{0.42\textwidth}
        \centering
        \vspace{-6mm}\includegraphics[width=1\linewidth]{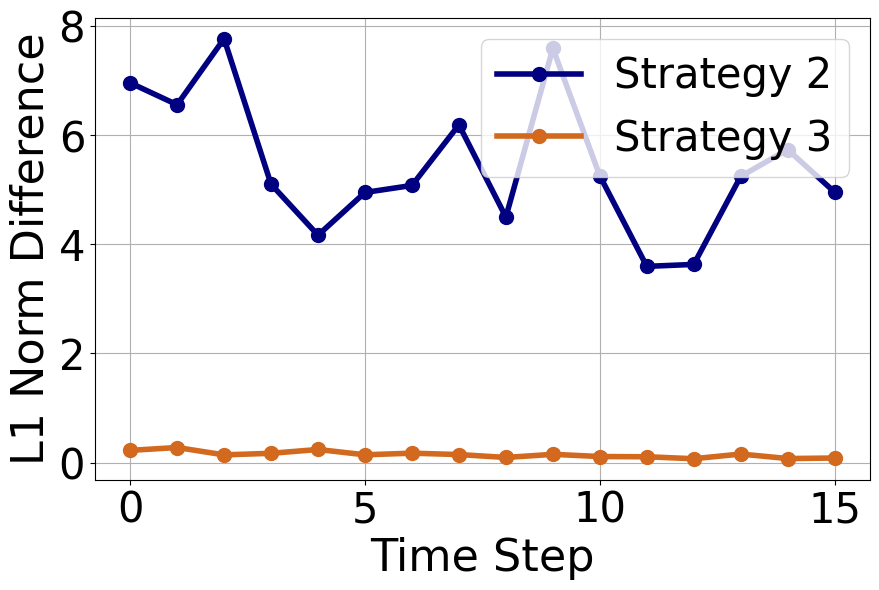}
    \captionof{figure}{L1 norm of differences between $\mW$ at two consecutive steps for Strategy 2 and Strategy 3}
    \label{fig:l1-norm}
    \end{minipage}
\end{figure}

\paragraph{A note on communication topology:} While the paper demonstrates a fully connected communication topology, our protocols can support sparser communication patterns, as long as the graph represented by $\mW$ is strongly connected (i.e., every vertex is reachable from every other vertex), though convergence will be slower.
We provide experimental results on the training time required for a ring topology to achieve the same perplexity level as a fully connected topology in Table~\ref{tab:ring-topology}. In high heterogeneity scenarios where each client is allocated a specific category, a ring topology results in cases where adjacent clients do not share any categories, leading to an \emph{oracle} trust weight of 0 and hence the NA entries. Despite this, our algorithm can still learn, converging to a perplexity approximately one point higher than reported for the fully connected case.

\begin{table}[t!]
\begin{center}
\begin{tabular}{l l | c c c } 
 \toprule
 \multirow{2}{*}{\textbf{Heterogeneity}} & \multirow{2}{*}{\textbf{Method}} & \multicolumn{3}{c}{\textbf{Datasets}} \\[0.2ex] \cline{3-5} \\[-0.8em]
  &  & AG News & Multi-Wiki & Codes-Wiki  \\ [0.2ex]
 \midrule
  \multirow{6}{*}{Low} & Local & $30.17(0.17)$ & $40.00(0.33)$ & $19.57(0.23)$ \\[0.2ex]
  & FedAvg & $31.66(0.20)$ & $52.75(0.57)$ & $17.53(0.19)$ \\[0.2ex]
  & FedAvg +FT & $32.25 (0.20)$ & $48.02 (0.25)$ & $20.17 (0.33)$ \\[0.2ex] \arrayrulecolor{black!30}\cmidrule{2-5}

  & Strategy 1 & $31.43(0.35)$ & $45.59(0.66)$ & $17.57(0.21)$ \\[0.2ex]
  & Strategy 2 & $\textbf{29.75(0.23)}$ & $36.93(0.17)$ & $17.61(0.40)$ \\[0.2ex]
  & Strategy 3 & 29.81(0.13)$^\star$ & $\textbf{36.70(0.23)}$ & $\textbf{17.35(0.18)}$ \\[0.2ex]
  \arrayrulecolor{black!30}\cmidrule{2-5}
  & \emph{Oracle} & $\mathit{29.56(0.21)}$ & $\mathit{39.55(0.19)}$ & $\mathit{17.42(0.19)}$ \\
  \arrayrulecolor{black}\midrule
  \multirow{6}{*}{High} & Local & $28.67(0.13)$ & $40.24(0.25)$ & $17.56(0.08)$ \\[0.2ex]
  & FedAvg & $32.08(0.13)$ & $53.23(0.51)$ & $16.68(0.06)$ \\[0.2ex]
& FedAvg +FT & $33.74(0.17)$ & $	48.07(0.18)$ & $18.43(0.18)$ \\[0.2ex]
\arrayrulecolor{black!30}\cmidrule{2-5}
  & Strategy 1 & $31.93(0.86)$ & $49.34(2.46)$ & $16.84(0.05)$ \\[0.2ex]
  & Strategy 2 & $\textbf{28.29(0.06)}$ & $37.20(0.20)$ & $16.22(0.17)$ \\[0.2ex]
  & Strategy 3 & 28.72(0.27)$^\star$ & $\textbf{36.92(0.16)}$ & $\textbf{16.23(0.12)}$ \\[0.2ex]
  \arrayrulecolor{black!30}\cmidrule{2-5}
  & \emph{Oracle} & $\mathit{28.08(0.11)}$ & $\mathit{35.96(0.24)}$ & $\mathit{16.20(0.05)}$ \\[0.2ex]
 
 \arrayrulecolor{black}\bottomrule
\end{tabular}
\end{center}
\caption{Test perplexities (standard deviation) of our proposed strategies and baseline methods (lower is better). Strategy 1: weights similarity-based; strategy 2: validation performance-based, strategy 3: prediction similarity-based. $\star$ denotes that the used public dataset is a synthetic dataset we constructed using ChatGPT.}
\label{table:res_summary}
\end{table}

\subsection{Trust matrix}
\begin{figure*}[t!]
    \centering
    \begin{subfigure}[t]{0.45\textwidth}
        \centering
        \includegraphics[width=\textwidth]{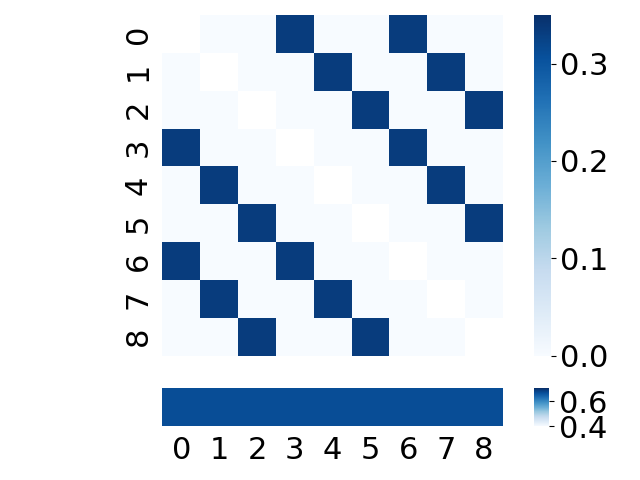}
        \caption{Oracle}
    \end{subfigure}%
    ~ 
    \begin{subfigure}[t]{0.45\textwidth}
        \centering
        \includegraphics[width=\textwidth]{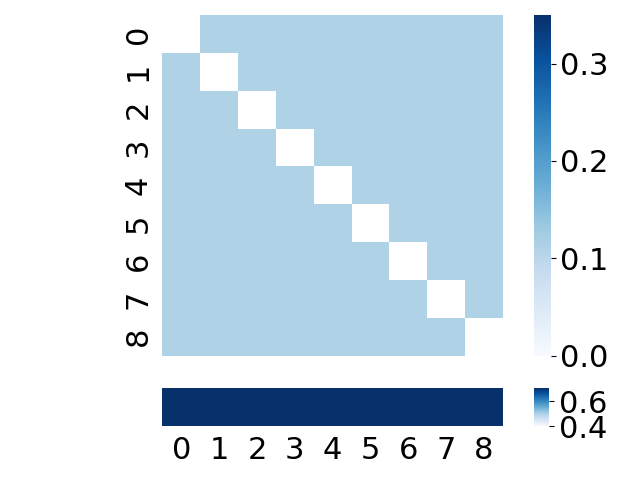}
        \caption{Strategy 1}
    \end{subfigure}
    \caption{Oracle trust matrix versus learned trust matrix using strategy 1 when users are allocated with Multilingual Wikipedia datasets. The diagonal entries are masked out and the trust is measured when the training ends. }
    \label{fig: trust-mat-1}
\end{figure*}

\begin{figure*}[t!]
    \centering
    \begin{subfigure}[t]{0.45\textwidth}
        \centering
        \includegraphics[width=\textwidth]{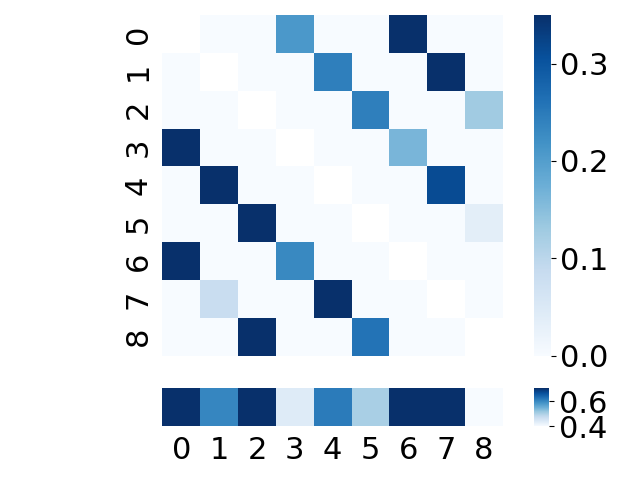}
        \caption{Strategy 2}
    \end{subfigure}%
    ~ 
    \begin{subfigure}[t]{0.45\textwidth}
        \centering
        \includegraphics[width=\textwidth]{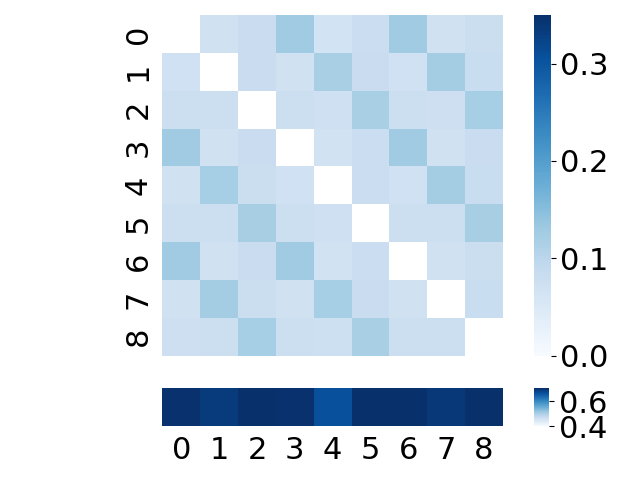}
        \caption{Strategy 3}
    \end{subfigure}
    \caption{Learned trust matrix using strategies 2 and 3 when users are allocated with Multilingual Wikipedia datasets. The diagonal entries are masked out and the trust is measured when the training ends. }
    \label{fig: trust-mat-2}
\end{figure*}

We compare our learned trust matrices with the \emph{oracle} trust matrix in Figures~\ref{fig: trust-mat-1} and~\ref{fig: trust-mat-2}. It is evident that strategies 2 and 3 uncover similar collaboration patterns as suggested by the \emph{oracle} matrix, whereas strategy 1 falls short in this regard. Specifically, strategy 1 assigns nearly identical trust weights to all other users, indicating nearly identical LoRA weights learned at convergence. This strange behavior is further investigated in Appendix~\ref{append: further investigation of strategy 1}: even without communication, the LoRA weights across users are almost equally different, suggesting that \emph{LoRA weights are not informative in collaborator selection when cosine similarity is used as the distance metric}.

The success of strategies 2 and 3 underscores the advantage of utilizing predictions to identify collaborators. It is interesting how the distinctions between different users' models become more pronounced after a forward pass. Such a phenomenon was not previously observed in the vision domain. We further noticed that trust matrices in strategy 3 are more stable across communication rounds compared to strategy 2, with fewer abrupt changes in values. As shown in Figure~\ref{fig:l1-norm}, the
$l_1$ norm of the differences between $\mW$s at consecutive steps fluctuates significantly for Strategy 2.  It is speculated that the more stable trust matrix in strategy 3 contributes to stabilizing the learning process, leading to better results.

\section{Conclusions}
Drawing from insights in collaborative learning communities, we propose 3 collaboration protocols tailored for jointly on-device LLM fine-tuning. To our best knowledge, this is the first work demonstrating that collaboration can be leveraged to improve personalization performance in language modeling. Our methods can efficiently address challenges encountered in personalized LLMs, such as limited availability of local data and data heterogeneity stemming from user characteristics.

Remarkably, \emph{predictions are more informative than model weights in identifying collaborators within the language domain}. Our Strategy 3, the predictions-based protocol, demonstrates promising practical applications for on-device deployment. It achieves superior performance while managing to maintain reasonable communication and computation costs.

\section{Discussions and Future Work}

In this study, we examined resource-constrained scenarios of collaborative LLM fine-tuning. While the investigation primarily delves into data heterogeneity, the exploration of model and resource diversities is also contemplated. For instance, in scenarios where users possess disparate capacities, the feasibility of devising a collaborative protocol to enhance model outcomes is questioned. \cite{cho2024heterogeneous} have developed collaboration protocols tailored for circumstances wherein users exhibit varying levels of support for LoRA ranks. An intriguing avenue for further exploration involves elucidating the efficacy of collaborative approaches when users are equipped with distinct numbers of LoRA modules corresponding to their capacities, and how such collaborative endeavors can yield mutual benefits.

Furthermore, within the scope of this research, particular emphasis is placed on unsupervised fine-tuning. We only evaluated Next Token Prediction capabilities. However, the prospect of supervised fine-tuning, i.e. instruction tuning, aimed at facilitating generalization to unseen tasks, is worth further investigation. An intriguing question arises: Can users enhance their generalization capabilities through collaboration?

Our protocols are centered around \emph{trust}. While the efficacy of trust in identifying collaborators has been thoroughly explored, there is potential for further investigation into its role in identifying adversarial users, which is one of the major challenges in Federated LLMs~\citep{chen2023federated}.

\newpage
\subsubsection*{Acknowledgments}
We thank Soumajit Majumder and Bettina Messmer for their helpful review and all the anonymous reviewers for their constructive feedback. We acknowledge funding from Huawei Cloud Intelligent Cloud Technologies Initiative, Google Research Collaborations, and from the Swiss National Science Foundation (SNSF) grant number 200020\_200342.

\bibliography{colm2024_conference}
\bibliographystyle{colm2024_conference}

\newpage
\appendix
\section{Appendix}

\subsection{Communication cost experiments}
\label{tab: communication costs in mb}

Tables~\ref{table:communication_complexity} is a more detailed analysis of the communication costs within each communication round. In this specific case, communicating logits is the most costly. However, the cost can be largely reduced (to $0.014\%$ of the original size) via sparse encoding techniques.

Note that different communicated units scale with different quantities. When the model size is larger, for example in billions, weights and weight updates can be larger than logits. In this case, logits can even emerge as a preferable communication unit over weights or gradients without compression.

\begin{table}[h]
\begin{center}
\begin{tabular}{p{5.5cm} | p{3.5cm} | p{3.5cm}}
\toprule
\bf Communicated Elements & \bf Communication Costs & \bf Measured Costs (MB) \\
\midrule
LoRA weights ($\vtheta_i$) & $\mathcal{O}(Ldk)$ & 2.359\\
LoRA weight updates ($\Delta \vtheta_i$) & $\mathcal{O}(Ldk)$ & 2.359\\
Logits ($f_{\vtheta_i}(\mX_S)$) & $\mathcal{O}(V)$ & 40.206\\
Sparse Top-K Logits ($f_{\vtheta_i}(\mX_S)_{COO}$) & $\mathcal{O}(K)$  & 0.560\\
\bottomrule
\end{tabular}
\end{center}
\caption{A comparison of communication complexity across the different shared units. $L$ denotes the number of layers, $d$ denotes the embedding size, $k$ is the LoRA rank, $V$ denotes the vocabulary size and $K$ denotes the number of logits kept for communication. $COO$ denotes the sparse encoding used, \textit{Coordinate list}. }
\label{table:communication_complexity}
\end{table}

\subsection{Hyper-parameters details}
\label{append :hyperparameters}

\begin{figure*}[t!]
    \centering
    \begin{subfigure}[t]{0.65\textwidth}
        \centering
        \includegraphics[width=\textwidth]{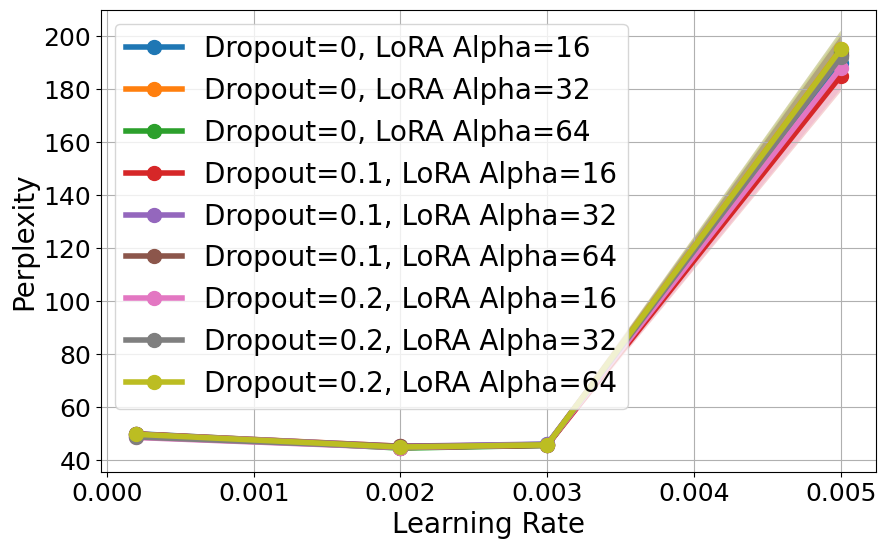}
        \caption{With respect to learning rate}
    \end{subfigure}
    ~ 
    \begin{subfigure}[t]{0.65\textwidth}
        \centering
        \includegraphics[width=\textwidth]{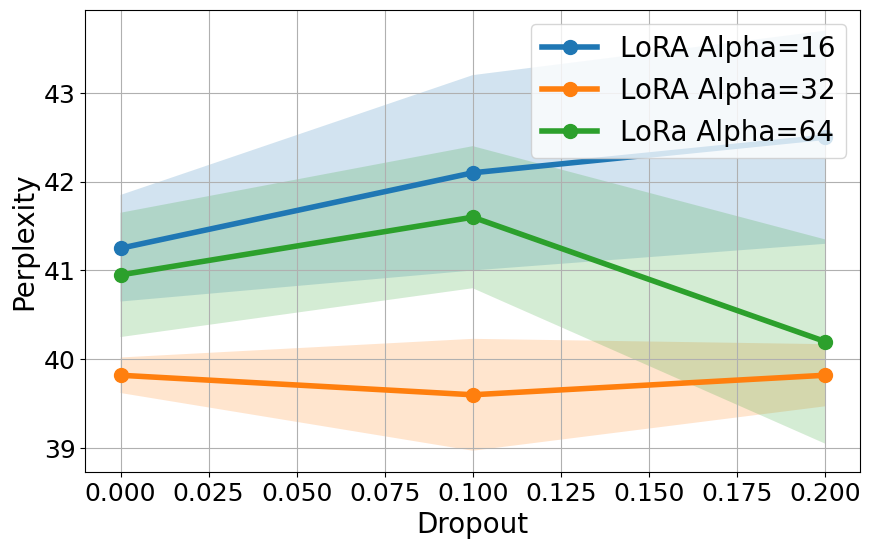}
        \caption{With respect to LoRA dropout}
    \end{subfigure}
    ~ 
    \begin{subfigure}[t]{0.65\textwidth}
        \centering
        \includegraphics[width=\textwidth]{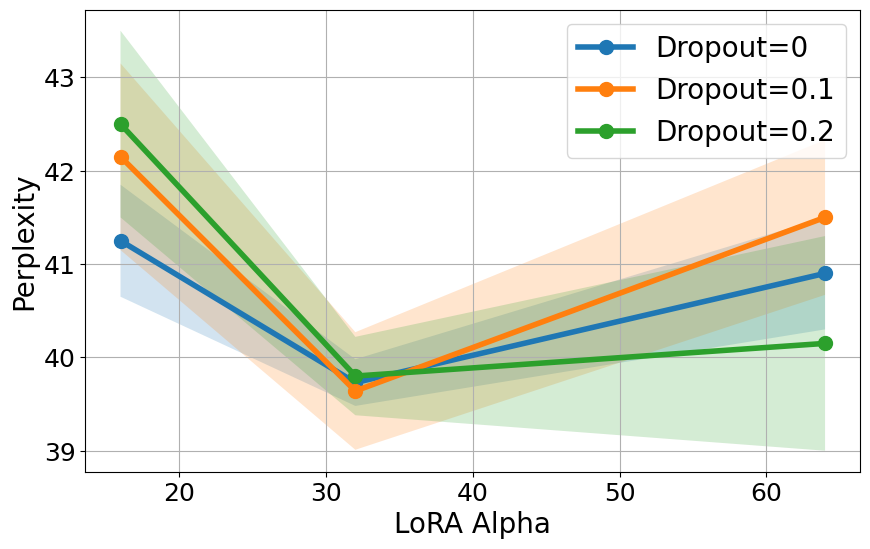}
        \caption{With respect to LoRA alpha}
    \end{subfigure}
    \caption{Hyperparameter tuning on learning rate, dropout, and LoRA alpha. LoRA dropout and LoRA alpha have a very small impact on the perplexity, the most important hyperparameter is the learning rate.}
    \label{fig:hyper_param_tuning}
\end{figure*}

The choice of our hyper-parameters is influenced by resource constraints, primarily due to memory limitations (training conducted on a single NVIDIA A100 GPU with 40GB memory) and time constraints, as we needed to fine-tune up to 9 large language models (LLMs). Through empirical observation, we found that a batch size of 50 with 4 accumulation steps and a context length of 512 performed well within memory constraints. Additionally, while increasing the LoRA rank slightly improved test perplexity, the associated increase in training time rendered it uninteresting for our experiments. Therefore, we opted for LoRA rank 4. Three hyper-parameters remain to be tuned: learning rate, LoRA alpha, and LoRA dropout. In Figure~\ref{fig:hyper_param_tuning}, the best perplexity was achieved with a learning rate of 0.002, LoRA dropout of 0.1, and LoRA alpha of 32.

\subsection{Additional Ablations}
\label{append: abaltions}

\subsubsection{Impact of local dataset size}
We increase the number of training tokens by 100 times, resulting in 84 million training tokens per client in the multilingual setup. The ranking of all methods remained consistent with what we report in Table~\ref{table:res_summary}.

\begin{table}[h!]
    \centering
    \begin{tabular}{l|c|c|c|c|c|c}
    \toprule
         Method & Local & FedAvg & Strategy 1 & Strategy 2 & Strategy 3 & Oracle \\
          \midrule
         val ppl & 37.80(0.15) & 56.69(0.35) & 48.47(2.55) & 36.17(0.30) & \textbf{35.63(0.15)} & 35.59 (0.16) \\
         \bottomrule
    \end{tabular}
    \caption{Comparison of methods on Multi-Wiki dataset}
    \label{app: tab-local-dataset-size}
\end{table}
\subsubsection{Impact of LoRA ranks}
In the main body, we restrict the LoRA rank to 4 to maintain a low computational budget. We conduct an ablation study on the LoRA ranks and present the results in Table~\ref{app: tab-lora-rank}. The rankings of the methods are consistent with our previous findings, with strategy 3 being the best method. Additionally, as expected, a higher LoRA rank results in lower test perplexity.
\begin{table}[h!]
    \centering
    \begin{tabular}{l|c|c|c|c}
    \toprule
   LoRA Ranks      &  4 & 8 &  16 &  32 \\
\midrule
 Local & 40.24(0.25) & 33.15(0.15) & 29.14(0.11) & 27.41(0.11) \\
 FedAvg & 53.23(0.51) & 48.84(0.34) & 42.68(0.06) & 40.12(0.25) \\
Strategy 1 & 49.34(2.46)& 48.37(1.26) & 40.50(0.11) & 34.65(5.73) \\
Strategy 2 & 37.20(0.20) & 30.41(0.13) & 26.36(0.12) & 24.37(0.20) \\
 Strategy 3  & \bf{36.92(0.12)} & \bf 30.25(0.19) & \bf 26.30(0.11) & \bf 24.39(0.02) \\
   \arrayrulecolor{black!30}\midrule
Oracle &  35.96(0.24) & 30.06(0.16)& 26.16(0.07) & 24.43(0.10)\\
\arrayrulecolor{black!100}\bottomrule
    \end{tabular}
    \caption{Comparison of different methods with various LoRA ranks}
    \label{app: tab-lora-rank}
\end{table}

\subsubsection{FedAvg with Local Fine-Tuning}
To understand better the relationship between the performance of FedAvg + Local Fine-Tuning and the relative size of the local fine-tuning dataset, we spare different percentages of local data for fine-tuning purposes, and the results are presented in Table~\ref{app:table:res_summary}. With more local fine-tuning data, the performance gets closer to Local training without collaboration. 

\begin{table}[h!]
\begin{center}
\begin{tabular}{l l | c c c } 
 \toprule
 \multirow{2}{*}{\textbf{Heterogeneity}} & \multirow{2}{*}{\textbf{Method}} & \multicolumn{3}{c}{\textbf{Datasets}} \\[0.2ex] \cline{3-5} \\[-0.8em]
  &  & AG News & Multi-Wiki & Codes-Wiki  \\ [0.2ex]
 \midrule
  \multirow{6}{*}{Low} & Local & $30.17(0.17)$ & $40.00(0.33)$ & $19.57(0.23)$ \\[0.2ex]
  & FedAvg & $31.66(0.20)$ & $52.75(0.57)$ & $17.53(0.19)$ \\[0.2ex]
  & FedAvg +FT (10\%) &  $32.25 (0.20)$ & $48.02 (0.25)$ & $20.17 (0.33)$ \\[0.2ex] 
   & FedAvg +FT (25\%) & $31.29 (0.15)$ & $43.74 (0.24)$ & $20.10 (0.25)$ \\
  [0.2ex] 
    & FedAvg +FT (50\%) & $30.03 (0.23)$ & $39.79(0.21)$ & $19.59(0.26)$ \\[0.2ex] \arrayrulecolor{black!30}\cmidrule{2-5}

  & Strategy 1 & $31.43(0.35)$ & $45.59(0.66)$ & $17.57(0.21)$ \\[0.2ex]
  & Strategy 2 & $\textbf{29.75(0.23)}$ & $36.93(0.17)$ & $17.61(0.40)$ \\[0.2ex]
  & Strategy 3 & 29.81(0.13)$^\star$ & $\textbf{36.70(0.23)}$ & $\textbf{17.35(0.18)}$ \\[0.2ex]
  \arrayrulecolor{black!30}\cmidrule{2-5}
  & \emph{Oracle} & $\mathit{29.56(0.21)}$ & $\mathit{39.55(0.19)}$ & $\mathit{17.42(0.19)}$ \\
  \arrayrulecolor{black}\midrule
  \multirow{6}{*}{High} & Local & $28.67(0.13)$ & $40.24(0.25)$ & $17.56(0.08)$ \\[0.2ex]
  & FedAvg & $32.08(0.13)$ & $53.23(0.51)$ & $16.68(0.06)$ \\[0.2ex]
& FedAvg +FT (10\%) & $33.74(0.17)$ & $	48.07(0.18)$ & $18.43(0.18)$ \\[0.2ex]

& FedAvg +FT (25\%) & 
$32.11(0.16)$ & $	43.68(0.13)$ & $18.15(0.11)$ \\[0.2ex]

& FedAvg +FT (50\%) & $29.84(0.09)$ & $	40.14(0.19)$ & $17.88(0.16)$ \\[0.2ex]

\arrayrulecolor{black!30}\cmidrule{2-5}
  & Strategy 1 & $31.93(0.86)$ & $49.34(2.46)$ & $16.84(0.05)$ \\[0.2ex]
  & Strategy 2 & $\textbf{28.29(0.06)}$ & $37.20(0.20)$ & $16.22(0.17)$ \\[0.2ex]
  & Strategy 3 & 28.72(0.27)$^\star$ & $\textbf{36.92(0.16)}$ & $\textbf{16.23(0.12)}$ \\[0.2ex]
  \arrayrulecolor{black!30}\cmidrule{2-5}
  & \emph{Oracle} & $\mathit{28.08(0.11)}$ & $\mathit{35.96(0.24)}$ & $\mathit{16.20(0.05)}$ \\[0.2ex]
 
 \arrayrulecolor{black}\bottomrule
\end{tabular}
\end{center}
\caption{(Extended) Test perplexities (standard deviation) of our proposed strategies and baseline methods (lower is better). Strategy 1: weights similarity-based; strategy 2: validation performance-based, strategy 3: prediction similarity-based. $\star$ denotes that the used public dataset is a synthetic dataset we constructed using ChatGPT.}
\label{app:table:res_summary}
\end{table}

\subsection{Weight similarity based trust further analysis}
\label{append: further investigation of strategy 1}
We offer a comprehensive analysis to elucidate why Strategy 1 underperforms the other two strategies. The convergent weight similarity-based trust matrix resembling a uniform matrix is noteworthy. This occurrence is peculiar, as we would anticipate heterogeneity due to the varying distributions of users' data.

\subsubsection{Strategy 1 trust matrix converges to uniform even without communication}
\begin{figure*}[h!]
    \centering
    \begin{subfigure}[t]{0.32\textwidth}
        \centering
        \includegraphics[width=\textwidth]{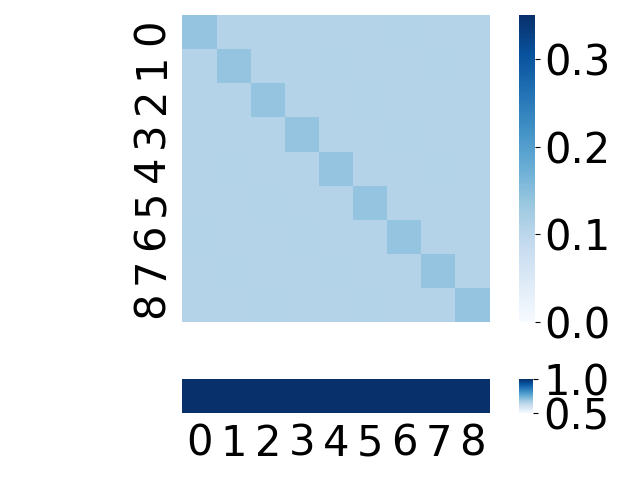}
        \caption{Round 1}
    \end{subfigure}%
    ~ 
    \begin{subfigure}[t]{0.32\textwidth}
        \centering
        \includegraphics[width=\textwidth]{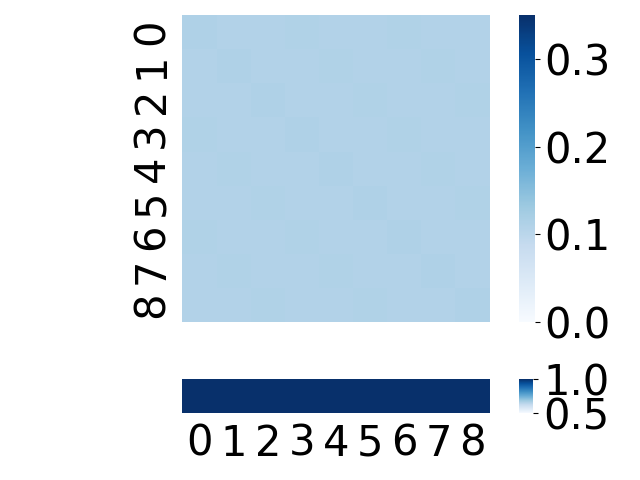}
        \caption{Round 2}
    \end{subfigure}
    ~ 
    \begin{subfigure}[t]{0.32\textwidth}
        \centering
        \includegraphics[width=\textwidth]{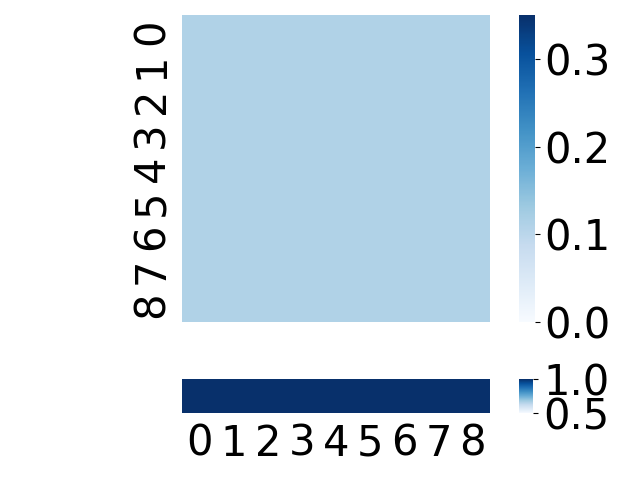}
        \caption{Last Round}
    \end{subfigure}
    \caption{Learned trust matrix using strategy 1 at different global rounds, without communication, when users are allocated with Multilingual Wikipedia datasets.}
    \label{fig:strategy- 1-trust-local-step-only}
\end{figure*}

\begin{figure*}[h!]
    \centering
    \begin{subfigure}[t]{0.32\textwidth}
        \centering
        \includegraphics[width=\textwidth]{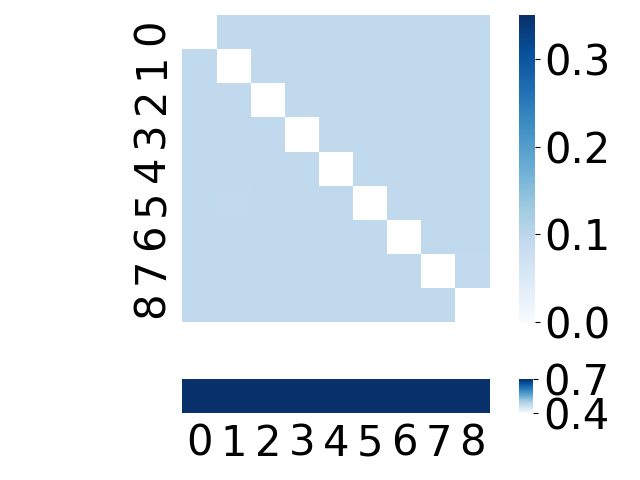}
        \caption{Strategy 1}
    \end{subfigure}%
    ~ 
    \begin{subfigure}[t]{0.32\textwidth}
        \centering
        \includegraphics[width=\textwidth]{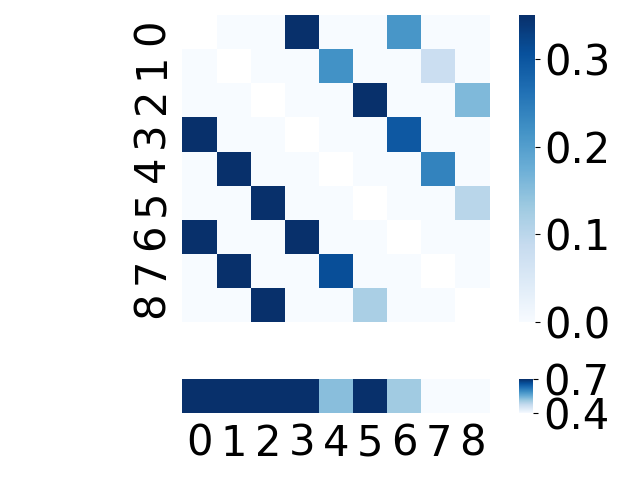}
        \caption{Strategy 2}
    \end{subfigure}
    ~ 
    \begin{subfigure}[t]{0.32\textwidth}
        \centering
        \includegraphics[width=\textwidth]{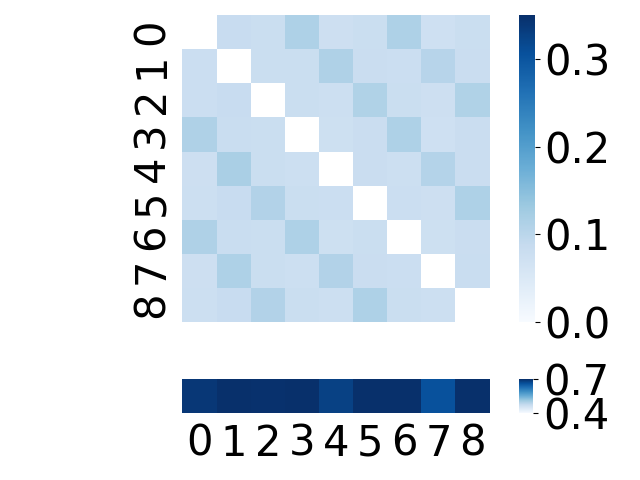}
        \caption{Strategy 3}
    \end{subfigure}
    \caption{Learned trust matrix using strategies 1, 2, and 3 when users are allocated with Multilingual Wikipedia datasets. No communication and aggregation are performed.}
    \label{fig:trust-mat-3-strategies}
\end{figure*}

We let each user perform local fine-tuning for 500 iterations ($\sim50$ epochs). Figure~\ref{fig:strategy- 1-trust-local-step-only} illustrates learned trust matrices using strategy 1 in different global rounds. Note that here, we use one global round to represent every 25 iterations, \emph{without} any actual communication between users. Figure~\ref{fig:trust-mat-3-strategies} shows the learned trust matrices at convergence using the three different strategies.

Even without communication, the learned trust matrix using Strategy 1 becomes uniform. This indicates that the users learned almost identical LoRA weights.

\subsubsection{Strategy 1 learns the collaboration pattern, but at a tiny scale}

\begin{figure*}[h!]
    \centering
    \begin{subfigure}[t]{0.32\textwidth}
        \centering
        \includegraphics[width=\textwidth]{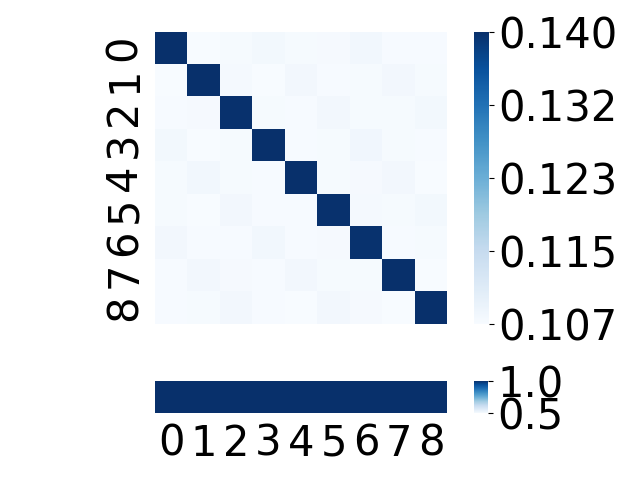}
        \caption{Round 1}
    \end{subfigure}%
    ~ 
    \begin{subfigure}[t]{0.32\textwidth}
        \centering
        \includegraphics[width=\textwidth]{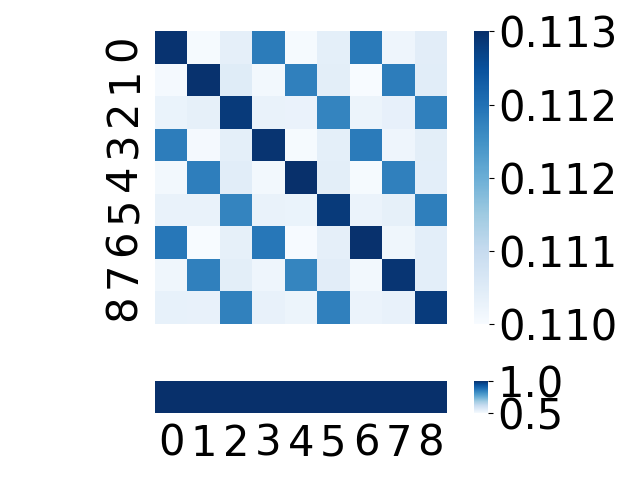}
        \caption{Round 2}
    \end{subfigure}
    ~ 
    \begin{subfigure}[t]{0.32\textwidth}
        \centering
        \includegraphics[width=\textwidth]{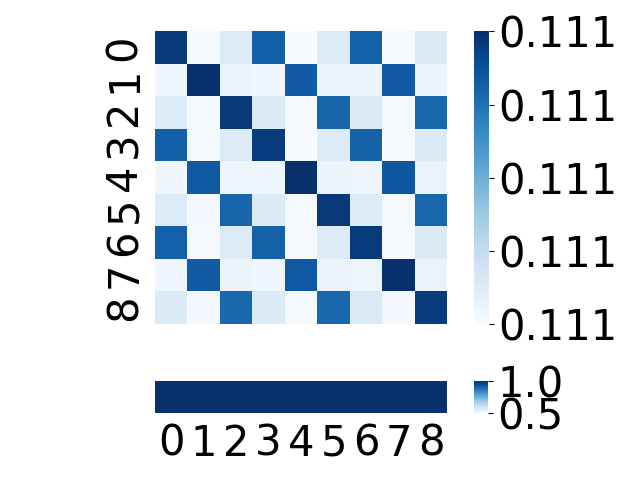}
        \caption{Last Round}
    \end{subfigure}
    \caption{Learned trust matrix using strategy 1 at different global rounds, without communication, when users are allocated with Multilingual Wikipedia datasets. Displayed with a smaller scale.}
    \label{fig:strategy- 1-trust-local-step-only-tiny-scale}
\end{figure*}

We dive into this bizarre observation.  In Figure~\ref{fig:strategy- 1-trust-local-step-only-tiny-scale}, we plot out the exact same trust matrices as in Figure~\ref{fig:strategy- 1-trust-local-step-only}, but present with a much smaller scale. This visualization indicates that Strategy 1 indeed can identify users who are more helpful to collaborate with, but the distinction between more and less helpful collaborators is not pronounced.

\end{document}